\documentclass[sigconf]{acmart}

\AtBeginDocument{%
  \providecommand\BibTeX{{%
    \normalfont B\kern-0.5em{\scshape i\kern-0.25em b}\kern-0.8em\TeX}}}

\setcopyright{acmlicensed}
\copyrightyear{2018}
\acmYear{2018}
\acmDOI{XXXXXXX.XXXXXXX}

\acmConference[MM'24]{Make sure to enter the correct
  conference title from your rights confirmation email}{October 28 - November 1,
  2024}{Melbourne, Australia.}
\acmISBN{978-1-4503-XXXX-X/18/06}

\usepackage{bm}
\usepackage{bbm}
\usepackage{mathrsfs}
\usepackage{multirow}
\usepackage{array}
\usepackage{wasysym}
\usepackage{booktabs}
\usepackage{algorithm}
\usepackage{algorithmic}

\usepackage{subfigure}

\begin{document}

\title{CLIP-Driven Cloth-Agnostic Feature Learning for Cloth-Changing Person Re-Identification}


\author{Shuang Li}
\affiliation{%
  \institution{Chongqing University of Posts and Telecommunications}
  \city{Chongqing}
  \country{China}}
\email{shuangli936@gmail.com}

\author{Jiaxu Leng}
\affiliation{%
  \institution{Chongqing University of Posts and Telecommunications}
  \city{Chongqing}
  \country{China}}
\email{lengjx@cqupt.edu.cn}

\author{Guozhang Li}
\affiliation{%
  \institution{Xidian University}
  \city{Xian}
  \country{China}}
\email{liguozhang@stu.xidian.edu.cn}

\author{Ji Gan}
\affiliation{%
  \institution{Chongqing University of Posts and Telecommunications}
  \city{Chongqing}
  \country{China}}
\email{jigan@cqupt.edu.cn}

\author{Haosheng Chen}
\affiliation{%
  \institution{Chongqing University of Posts and Telecommunications}
  \city{Chongqing}
  \country{China}}
\email{haosheng.chen@hotmail.com}

\author{Xinbo Gao}
\affiliation{%
  \institution{Chongqing University of Posts and Telecommunications}
  \city{Chongqing}
  \country{China}}
\email{gaoxb@cqupt.edu.cn}
\renewcommand{\shortauthors}{author name and author name, et al.}

\begin{abstract}
Contrastive Language-Image Pre-Training (CLIP) has shown impressive performance in short-term Person Re-Identification (ReID) due to its ability to extract high-level semantic features of pedestrians, yet its direct application to Cloth-Changing Person Re-Identification (CC-ReID) faces challenges due to CLIP's image encoder overly focusing on clothes clues. To address this, we propose a novel framework called CLIP-Driven Cloth-Agnostic Feature Learning (CCAF) for CC-ReID. Accordingly, two modules were custom-designed: the Invariant Feature Prompting (IFP) and the Clothes Feature Minimization (CFM). These modules guide the model to extract cloth-agnostic features positively and attenuate clothes-related features negatively. Specifically, IFP is designed to extract fine-grained semantic features unrelated to clothes from the raw image, guided by the cloth-agnostic text prompts. This module first covers the clothes in the raw image at the pixel level to obtain the shielding image and then utilizes CLIP's knowledge to generate cloth-agnostic text prompts. Subsequently, it aligns the raw image-text and the raw image-shielding image in the feature space, emphasizing discriminative clues related to identity but unrelated to clothes. Furthermore, CFM is designed to examine and weaken the image encoder's ability to extract clothes features. It first generates text prompts corresponding to clothes pixels. Then, guided by these clothes text prompts, it iteratively examines and disentangles clothes features from pedestrian features, ultimately retaining inherent discriminative features. Extensive experiments have demonstrated the effectiveness of the proposed CCAF, achieving new state-of-the-art performance on several popular CC-ReID benchmarks without any additional inference time.
\end{abstract}




\maketitle

\section{Introduction}
Person Re-identification (ReID) \cite{li2018harmonious,sun2018beyond,hou2019interaction,zhu2020identity} aims to accurately recognize and retrieve pedestrians with the same identity across different cameras. This technology holds great promise in fields such as intelligent monitoring and intelligent security \cite{ye2021deep}. However, existing methods are largely confined to short-term ReID within limited time and space, assuming that pedestrians wear the same clothes \cite{sun2018beyond,li2018harmonious,hou2019interaction}. In reality, pedestrians with the same identity often wear different clothes across a broader range of time and space \cite{gu2022clothes,qian2020long}. Additionally, fugitives frequently change their clothes to evade capture. Therefore, directly applying short-term ReID methods to these scenarios will significantly decrease their performance, as short-term ReID methods overly rely on clothes information (such as pattern, color, etc.) as discriminative clues for pedestrians.

\begin{figure}[t!]
\centering
\includegraphics[width=8cm,keepaspectratio=true]{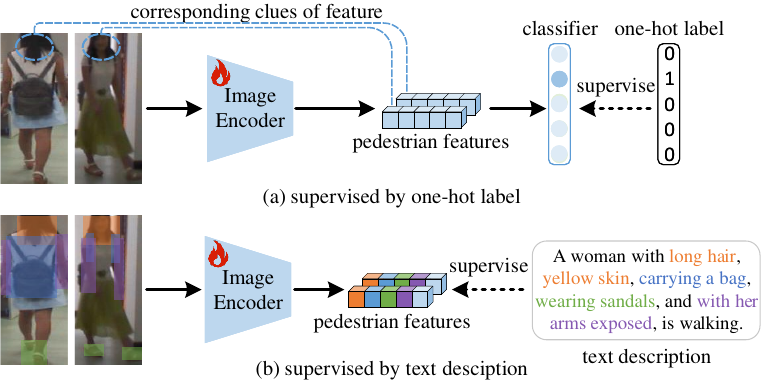}
\caption{Compared to the one-hot label, the text description can provide more specific semantic descriptions. Therefore, training with text descriptions enables the model to focus more on discriminative clues \cite{yu2023tf}.  }
\label{intro1}
\end{figure}

To mitigate the impact of clothes changes on ReID, the CC-ReID was proposed to enhance the performance of the ReID system under varying clothes conditions.
Some scholars have proposed many excellent methods, some of which focus on extracting biometric features, including contour sketches \cite{yang2019person}, body shape \cite{chen2021learning,hong2021fine,wang2022co}, face \cite{wan2020person}, and gait \cite{jin2022cloth}, while others concentrate on using disentanglement methods to separate clothes features from pedestrian features \cite{xu2021adversarial,cui2023dcr,yang2023good}.
These methods commonly follow a structure of ``Image encoder + pre-training weights from ImageNet + one-hot label’’. 
Here, the image encoder is initialized with weights obtained from training on ImageNet, 
and the ReID model training is typically supervised by the one-hot label. 
As shown in Fig. \ref{intro1}, compared to the text description, the one-hot label contains relatively less information, limiting the ability of the image encoder to learn fine-grained semantic features.

\begin{figure}[t!]
\centering
\includegraphics[width=8cm,keepaspectratio=true]{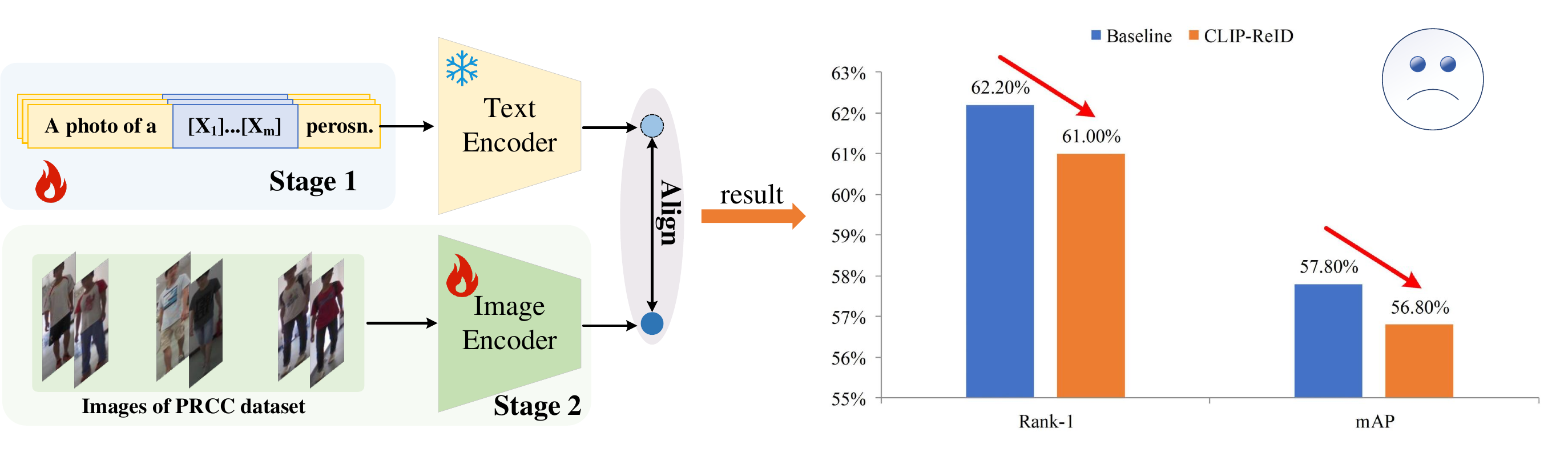}
\caption{An illustration of the application of CLIP-ReID to the PRCC dataset of CC-ReID and the experimental results obtained.  }
\label{intro}
\end{figure}

Recently, the introduction of Contrastive Language-Image Pre-Training (CLIP) \cite{radford2021learning} has offered a fresh approach to tackling this challenge, as it connects visual representations with high-level language descriptions through contrastive learning. In the field of person ReID, CLIP-ReID \cite{li2023clip}, an exceptional prompt learner, has pioneered the application of CLIP and achieved success in short-term person ReID. It addresses the issue of lacking image-text correspondence in person ReID by freezing image and text encoders and generating text prompts associated with person identities, as shown in Fig.\ref{intro}.
To explore the application of CLIP in CC-ReID, we hired CLIP-ReID and directly applied it to CC-ReID.
However, as shown in Fig.\ref{intro}, its performance exhibited significant degradation. 
Intuitively, the introduction of prompts led to a degradation in performance, indicating that the generated prompts did not describe identity-related clues but rather tended to describe clothes-related clues (such as color, and patterns of clothes). 
The capture of these clothes clues is obviously advantageous in short-term person ReID, as clothes itself is a discriminative clue. 
However, in CC-ReID, these clues turned into noise. Therefore, our primary task is to explore \textbf{how to generate cloth-agnostic text prompts and how to utilize these prompts to learn cloth-agnostic features}. 

In this paper, we propose the CCAF framework, which has two objectives, i.e.,  (a) to positively guide the model to extract cloth-agnostic features; and (b) to negatively weaken the model's extraction of clothes-related features. For the first objective, it is crucial to obtain cloth-agnostic text prompts. Considering that the text prompts are generated based on the input pedestrian image, if there is no clothes clues in the input image, the generated text prompts will not describe clothes-related clues. This means that we can fine-tune the model to focus on cloth-agnostic visual features through the guidance of cloth-agnostic prompts. To achieve this, we first erase the pixels corresponding to clothes to obtain shielding images, thus generating cloth-agnostic text prompts.
Then we can guide the model to focus on cloth-agnostic clues based on the cloth-agnostic text prompts. Secondly, considering the inherent modality differences between text and image, we utilize cloth-agnostic text prompts to guide the model to extract cloth-agnostic visual features from shielding images. Additionally, we align the shielding images and raw images in the feature space to further strengthen the model's focus on cloth-agnostic discriminative clues.

To this end, we propose the Invariant Feature Prompting (IFP). Specifically, in the first stage of IFP, we employ SCHP \cite{li2020self} to parse pedestrian images, mask out clothes pixels, and obtain shielding images. Then we use the Generate Cloth-Agnostic Text Prompt (GCATP) to generate text prompts corresponding to the shielding images. In the second stage, there are two branches: the raw stream and the shielding stream. In the raw stream, we establish a correlation between the visual features of raw images and the text features of text prompts through Image-Text Alignment (I2T). Considering the inherent modality differences between text and image, we train the shielding stream to learn the visual features of the shielding images and transfer this capability to the raw stream branch through Image-Image Alignment (I2I). Compared to previous work, the proposed IFP encourages the model to directly extract high-level semantic features from raw images corresponding to cloth-agnostic text prompts, enhancing feature discriminability.

Although cloth-agnostic text prompts can guide the model in learning how to extract cloth-agnostic visual features from raw images, 
the lack of a mechanism to examine the absence of clothes-related clues in pedestrian features means that the pedestrian features may still contain noise related to clothes. Therefore, we also propose a novel approach to verify whether the model can still extract features related to clothes and mitigate this through fine-tuning.
 Specifically, leveraging the powerful capabilities of CLIP, we generate text prompts for the clothes images that contain only pixels of clothes. Guided by these prompts, we encourage the projection of clothes features from pedestrian features, thereby examining whether the pedestrian features contain clothes clues. Furthermore, by fine-tuning the image encoder, we encourage the generated pedestrian features to move away from clothes features. Ultimately, this approach iteratively weakens the model's reliance on clothes clues and strengthens its ability to extract features that are closely related to individual identities and independent of clothes, through a cycle of "examine-disentangle-examine".

To examine whether there are still clothes features in pedestrian features, we introduce the Clothes Feature Minimization (CFM). This module first uses the Generate Clothes Text Prompt (GCTP) to generate clothes text prompts for clothes images in the first stage, similar to GCATP. Then, the Feature Disentanglement (FD) employs a projection matrix to project clothes features from pedestrian features of the raw images based on the clothes text prompts, further pushing away the pedestrian features of raw images from the clothes features. This process iterates continuously, with CFM weakening the model's ability to extract clothes features from the opposite side, effectively complementing IFP.  Together, they address the core issue of CLIP focusing excessively on clothes clues.

 Our main contributions are summarized as follows:
\begin{itemize}
    \item The proposed CCAF successfully extracts cloth-agnostic features. To the best of our knowledge, this is the first exploration of applying CLIP's cross-modal knowledge to CC-ReID.
    \item We explored a simple yet effective method to generate cloth-agnostic text prompts and clothe text prompts, providing explicit supervision for fine-tuning the image encoder.
    \item Our proposed IFP and CFM enhance the learning of cloth-agnostic pedestrian features from both positive and negative aspects, guided by the generated cloth-agnostic and clothes text prompts.
    \item Extensive experimental results on PRCC, LTCC, VC-Clothes, and Deepchage datasets show the proposed CCAF achieves a new state-of-the-art performance.
\end{itemize}

\vspace{-0.21cm}
\section{Related Work}
\subsection{Cloth-changing Person Re-Identification}
For short-term ReID methods, clothes is often considered an effective discriminative clue. However, in CC-ReID, this thread may hinder pedestrian retrieval performance. This is because individuals with the same identity may change clothes on different occasions or periods. Therefore, in the CC-ReID task, we need to explore more robust and stable feature representation methods to address the challenges posed by clothes changes. To overcome this problem, researchers are exploring other types of features, such as body shape \cite{hong2021fine,chen2021learning,yang2019person}, and gait \cite{jin2022cloth} to assist in person ReID. In addition, some scholars have also attempted to separate clothes features through disentanglement to retain pedestrian features that are irrelevant to clothes \cite{xu2021adversarial,cui2023dcr,gu2022clothes}. Specifically, to focus on clothes-irrelevant biological cues, Yang et al. \cite{yang2019person} introduce the spatial polar transformation layer into the neural network to extract discriminative curve clues of body shape from contour sketch images. FSAM \cite{hong2021fine} proposes a dual-stream network framework that integrates body shape features into appearance features through mutual learning. Furthermore, 3DSL \cite{chen2021learning} directly extracts texture-insensitive 3D body shape features from 2D images to resist the interference of clothes changes by performing the auxiliary task of reconstructing 3D body from 2D images. To capture and extract gait features from a single image, GI-ReID \cite{jin2022cloth} proposed to reconstruct gait images before and after frames from a single pedestrian image and extract gait features, then integrate them with appearance features.
To disentangle clothes features and pedestrian features, CAL \cite{xu2021adversarial} proposes a Clothes-based Adversarial Loss to extract clothes-irrelevant features from original RGB images. AFD-Net \cite{gu2022clothes} achieves intra-class reconstruction and inter-class outfit swapping through adversarial learning to disentangle identity-related and identity-unrelated (clothes) features. Considering the instability of adversarial learning, DCR \cite{cui2023dcr} proposes a controllable way to achieve disentanglement by reconstructing human component regions. 
However, these methods can only extract limited features due to being supervised solely by one-hot labels, without fully utilizing richer text descriptions.

\vspace{-4pt}
\subsection{CLIP in Person Re-identification}
Vision Language Pre-training aims to establish a close relationship between language and images by training on large-scale image-text pair data. The representative Contrastive Language-Image Pre-Training (CLIP) \cite{radford2021learning} of Vision Language Pre-training has demonstrated strong capabilities in multiple downstream tasks, including semantic segmentation \cite{zhou2023zegclip}, image-text retrieval \cite{ma2022x}, and video caption \cite{tang2021clip4caption}. In the person ReID community, recent research has explored the application of CLIP in person ReID \cite{yan2023clip, jiang2023cross, li2023clip, he2023region, yu2023tf}, achieving significant success.
In various subtasks of ReID, Text-to-image Person Retrieval (TI-ReID) naturally possesses paired text-image data. Considering that the pre-training process of CLIP relies mainly on instance-level data, which may lack the ability to learn fine-grained features, CFine \cite{yan2023clip} explores multi-granularity pedestrian feature learning based on the introduction of the CLIP image encoder. To more comprehensively explore the potential of CLIP in TI-ReID, IRRA \cite{jiang2023cross} extensively introduces both the CLIP image encoder and text encoder and achieves implicit alignment of local features.
In image-based person ReID, directly applying CLIP technology to extract semantic features of pedestrians becomes particularly challenging due to the lack of corresponding text descriptions for pedestrian images. To tackle this challenge, CLIP-ReID \cite{li2023clip} innovatively proposes generating text prompts closely related to identities for pedestrian images by training a prompt learner. TF-ReID \cite{yu2023tf} directly replaces text features with CLIP memory to explore the application of CLIP in video-based person re-identification.
Recognizing the strong robustness of language to modality changes, CSDN \cite{yu2024clip} is the first to apply CLIP to the Visible-Infrared person ReID. By cleverly using language as a bridge, CSDN successfully establishes a match of high-level semantic features between infrared and visible images. However, the enormous potential of CLIP in advancing the learning of cloth-agnostic features for CC-ReID has not been fully explored. In this paper, we propose the CCAF framework, which effectively leverages CLIP's cross-modal knowledge to explore the generation of cloth-agnostic text prompts and clothes prompts, as well as learning cloth-agnostic semantic features from both positive and negative aspects based on these text prompts.

\begin{figure*}[t!]
\centering
\includegraphics[width=16.7cm,keepaspectratio=true]{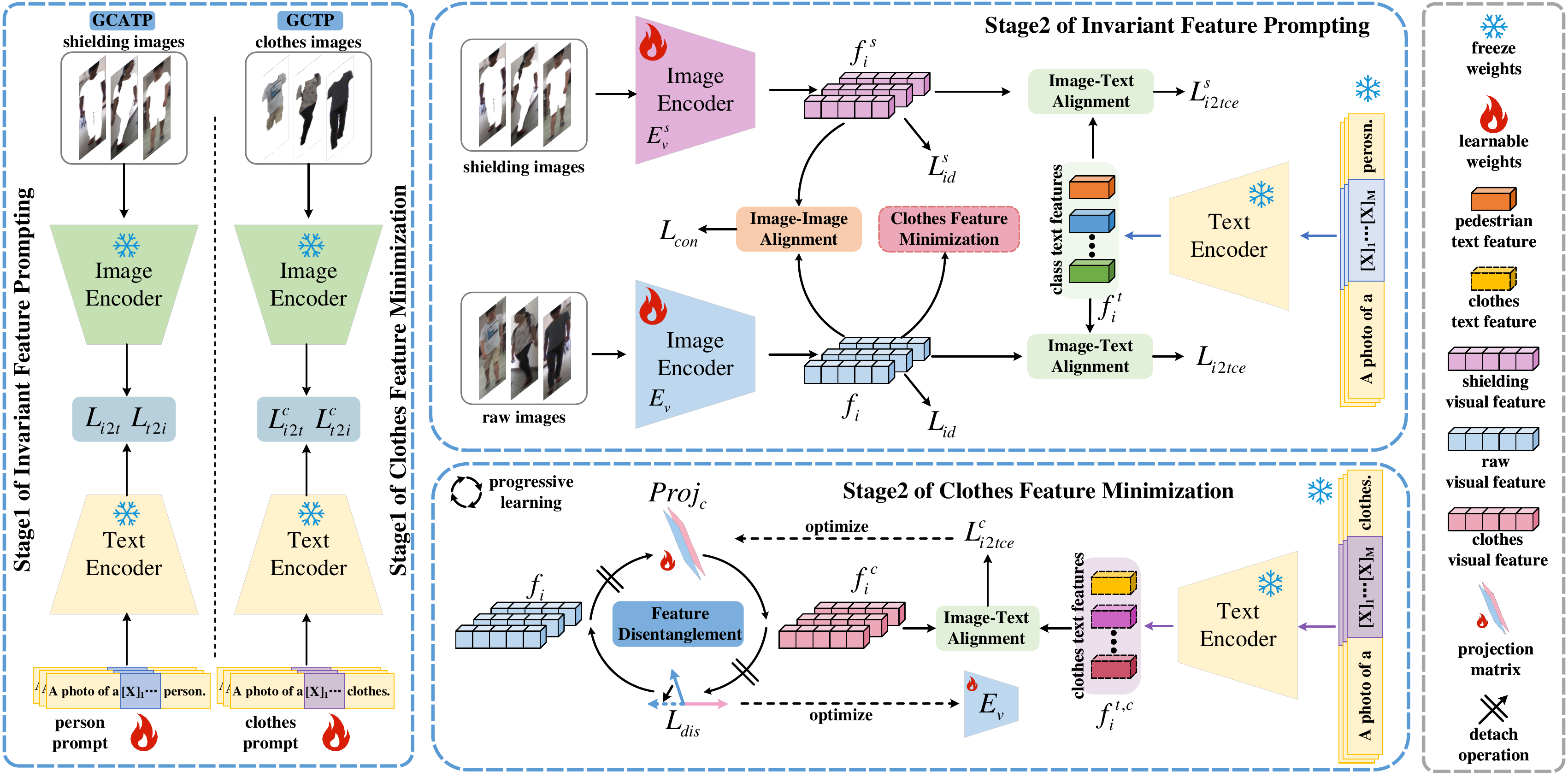}
\caption{The two-stage framework, CCAF, consists of Invariant Feature Prompting (IFP) and Clothes Feature Minimization (CFM). IFP learns cloth-agnostic features directly from raw images through Image-Text Alignment (I2T) and Image-Image Alignment (I2I). CFM examines whether pedestrian features contain clothes features and disentangles clothes features from pedestrian features.
}
\label{framework}
\end{figure*}

\section{Methodology}

\subsection{Preliminaries}\label{sec3.1}
Let $X=\{(x_{i}, y^{id}_{i},y^{c}_{i})|_{i=1}^{N}\}$ represent the training dataset, where $x_{i}$ represents the $i$-th image, $N$ represents the total number of pedestrian images, $y^{id}_{i}$  represents its corresponding identity label and $y^{c}_{i}$ represents its corresponding clothes label.
In existing CC-ReID research, a standard image encoder with pre-trained weight from ImageNet is typically employed to extract pedestrian features $f_i$ from the raw pedestrian images $x_i$.  To ensure the pedestrian features $f_i$ have identity discriminability, the cross-entropy loss $\mathcal{L}_{ce}$ \cite{sun2018beyond} and triplet loss loss $\mathcal{L}_{tri}$ \cite{hermans2017defense} are employed to constrain  $f_i$:
\begin{equation}
\begin{aligned}
\mathcal{L}_{ce}&=-\frac{1}{n_{b}}\sum_{i=1}^{n_{b}}q_{i}\log(\bm W_{id}(\bm f_{i})))
\end{aligned},
\end{equation}

\begin{equation}
\begin{aligned}
\mathcal{L}_{tri}=&\frac{1}{n_{b}}\sum_{i=1}^{n_{b}}[m+\|\bm f_{i}-\bm f_{i}^{p}\|_{2}-\|\bm f_{i}-\bm f_{i}^{n}\|_{2}]_{+}
\end{aligned},
\end{equation}
where $n_{b}$ represents the batch size, $q_{i}\in\mathbb{R}^{K \times 1}$ is a one-hot vector, and only the element at $y^{id}_{i}$ is $1$, $\bm W_{id}$ represents the identity classifier consisting of the BN layer and Liner layer followed by BNNeck \cite{luo2019bag},  $m$ is set to 0.3 empirically,  $f_{i}^{n}$ and $\bm f_{i}^{p}$ are the hard-negative and hard-positive samples of $\bm f_{i}$, respectively, $[\cdot]_{+}=max(\cdot,0)$ is the hinge loss. The overall optimization objective is defined as:
\begin{equation}
\begin{aligned}
\mathcal{L}_{id}=\mathcal{L}_{ce} + \mathcal{L}_{tri}
\end{aligned}.
\end{equation}
However, as discussed in the introduction, relying solely on one-hot labels for model training cannot fully exploit the semantic information relevant to pedestrian identity. To alleviate this limitation, in this paper, we replaced the traditional visual encoder with CLIP's visual encoder $E_v$ and used it as our baseline method. Based on this, we propose IFP and CFM to learn intrinsic high-level semantic features of pedestrians from both positive and negative aspects.

\subsection{Invariant Feature Prompting }\label{sec3.2}
The Invariant Feature Prompting (IFP) comprises three components: the Generate Cloth-Agnostic Text Prompt (GCATP), the Image-Text Alignment (I2T), and the Image-Image Alignment (I2I). GCATP is designed to generate cloth-agnostic text prompts. Subsequently, I2T facilitates the raw stream network $E_v$ in directly learning clothes-agnostic features from raw images. Finally, I2I aligns the output features of $E_v$ with the output features of the shielding stream network $E_v^s$, transferring the ability of the latter to the former.

\textbf{Generate Cloth-Agnostic Text Prompt}. 
To obtain the identity-specific text prompt of the pedestrian images, one straightforward approach is to directly apply the CLIP-ReID method to the pedestrian images. However, as illustrated in Fig. \ref{intro}, impure text prompts generated can misguide the image encoder, leading to an overemphasis on clothes clues. The main reason for this phenomenon is that CLIP's image encoder tends to focus on color and patterns, which are predominantly from clothes in pedestrian images. \textbf{If the input image lacks pixels of clothes, the text prompts generated only focus on discriminative clues rather than clothes-related clues}.

To achieve this goal, we employ a state-of-the-art human parsing model SCHP \cite{li2020self} to parse the input pedestrian image $x_{i}$ and obtain the human parsing result, which includes 20 categories such as background, hair, upper clothes, dress, coat, pants, skirt, face, left arm, and so on. We change the pixel values of clothes to 0 and the rest to 1, obtaining the masks $M^{p}_{i}$ of clothes-irrelevant areas. Then, we obtain the shielding images $x^{s}_{i}$ as follows:
\begin{equation}
\begin{aligned}
x^{s}_{i} =  & M^{p}_{i} \odot x_{i} + (I-M^{p}_{i})  \odot 255
\end{aligned},
\end{equation}
where $\odot$ denotes the Hadamard product. After eliminating the influence of clothes information at the pixel level, we introduced identity-related learnable tokens inspired by CLIP-ReID to learn the cloth-agnostic text prompt. 
Specifically, we designed an initial language description $T_{y_i^{id}}$ for the identity ${y_i^{id}}$ of the pedestrian: "A photo of a $[X]_1$ $[X]_2$ ... $[X]_M$ person", 
where $[X]_i$ represents the learnable tokens and $M$ represents the number of learnable tokens. As shown in Fig. \ref{framework}, in the first stage of training, we freeze CLIP's image encoder $E_{v}^{}$ and text encoder $E_{t}$ and then fed  $x^{s}_{i}$ and $T_{y_i^{id}}$ into the image encoder $E_{v}$ and text encoder $E_{t}$ to obtain image features $\bm f^{s}_{{i}} = E_{v}(x^{s}_{i}) $ and text features $\bm f^{t}_{{y_i^{id}}} = E_{t}(T_{y_i^{id}}) $. To ensure that the learnable $T_{y_i^{id}}$ can describe the identity information of pedestrians, we optimized the learnable $T_{y_i^{id}}$ through the image-to-text loss $\mathcal{L}_{i2t}$ and the text-to-image loss $\mathcal{L}_{t2i}$ as follows:
\begin{equation}
	\begin{aligned}
	\mathcal{L}_{i2t} =  - \frac{1}{{{n_b}}}\sum\limits_{i = 1}^{{n_b}}  {\log \frac{{\exp \left( {s\left( {\bm f_{i}^s,\bm f_{y_i^{id}}^{t}} \right)} \right)}}{{\sum\nolimits_{j = 1}^{{n_b}} {\exp \left( {s\left( {\bm f_i^s,\bm f_{y_j^{id}}^{t}} \right)} \right)} }}}   
	\end{aligned},
\end{equation}

\begin{equation}
	\begin{aligned}
	\mathcal{L}_{t2i} =   - \frac{1}{{{n_b}}}\sum\limits_{i = 1}^{{n_b}}  {\log \frac{{\exp \left( {s\left( {\bm f_{i}^s,\bm f_{y_i^{id}}^{t}} \right)} \right)}}{{\sum\nolimits_{j = 1}^{{n_b}} {\exp \left( {s\left( {\bm f_j^s,\bm f_{y_i^{id}}^{t}} \right)} \right)} }}}  
	\end{aligned},
\end{equation}
where $s(\cdot)$ represents the similarity between two feature tensors. 

\textbf{Image-Text Alignment}.
In the fine-tuning stage of the second stage, we froze the text encoder $E_{t}$ and $T_{y_i^{id}}$ and only trained the image encoder $E_v$. 
Specifically, we fed the prompts $T_{y_i^{id}}$ obtained in GCATP into the text encoder $E_{t}$ to obtain their corresponding text features $\bm f_{y_i^{id}}^{t} = E_t(T_{y_i^{id}})$ and fed $x_i$ into $E_v$  to obtain the raw feature $\bm f_i = E_v(x_i)$. Then, we employ loss $\mathcal{L}_{i2tce}$ to constrain the raw features $\bm f_i$ to be close to the corresponding text features $\bm f_{y_i^{id}}^{t}$ while being far from the text features of other identities:

\begin{equation}
	\begin{aligned}
	\mathcal{L}_{i2tce} =  - \frac{1}{{{n_b}}}\sum\limits_{i = 1}^{{n_b}}  {\log \frac{{\exp \left( {s\left( {\bm f_{i},\bm f_{y_i^{id}}^{t}} \right)} \right)}}{{\sum\nolimits_{y^{id} = 1}^{{K}} {\exp \left( {s\left( {\bm f_i,\bm f_{y^{id}}^{t}} \right)} \right)} }}}   
	\end{aligned},
\end{equation}
where $K$ is the total number of identities.

\textbf{Image-Image Alignment}.
Considering the inherent modality differences between text and image, it is challenging to ensure that $E_v$ can directly extract discriminative pedestrian clues unrelated to clothes from raw images solely through I2T. Therefore, we also designed the Image-Image Alignment (I2I) to ensure the semantic consistency between the features of the shielding images and the raw images, thereby transferring the knowledge learned by the shielding stream to the raw stream within the modality.
Specifically, we employ the image encoder $E_v^s$ to extract features $f_i^s = E_v^s(x_i^s)$. and then fine-tune $E_v^s$ through $\mathcal{L}_{id}^{s}$ and $\mathcal{L}_{i2tce}^{s}$ as follows: 
\begin{equation}
\begin{aligned}
\mathcal{L}_{id}^{s}=\mathcal{L}_{ce}^{s} + \mathcal{L}_{tri}^{s}
\end{aligned},
\end{equation}
\begin{equation}
	\begin{aligned}
	\mathcal{L}_{i2tce}^{s} =  - \frac{1}{{{n_b}}}\sum\limits_{i = 1}^{{n_b}}  {\log \frac{{\exp \left( {s\left( {\bm f_{i}^s,\bm f_{y_i^{id}}^{t}} \right)} \right)}}{{\sum\nolimits_{y^{id} = 1}^{{K}} {\exp \left( {s\left( {\bm f_i^s,\bm f_{y^{id}}^{t}} \right)} \right)} }}}   
	\end{aligned},
\end{equation}
where $\mathcal{L}_{ce}^{s}$ and $\mathcal{L}_{tri}^{s}$ represent cross entropy loss and triplet loss, respectively. And $\bm f_{y^{id}}^{t} = E_t(T_{y^{id}})$ represent the feature of text prompt $T_{y^{id}}$.
This allows $E_v^s$ to extract discriminative features that are independent of clothes from the shielding images, as the shielding images do not contain these clothes' information at the pixel level.
To transfer the ability of $E_v^s$ to $E_v$ and further enhance the ability of $E_v$ to extract the cloth-agnostic features, we adopt a mutual learning manner \cite{hong2021fine,jin2022cloth}  to ensure semantic consistency between $f_i^s$ and $f_i$:
\begin{equation}
	\begin{aligned}
	\mathcal{L}_{con} =  \frac{1}{{{P}}}\sum\limits_{i = 1}^{{P}} \| \bm c_{i} -\bm c_{i}^s\|_2^2 \\
    \bm c_{i}^s = \frac{1}{{{K_p}}}\sum\limits_{i = 1}^{{K_p}}\bm f_{i}^s, 
    \bm c_{i} = \frac{1}{{{K_p}}}\sum\limits_{i = 1}^{{K_p}}\bm f_{i}
	\end{aligned},
\end{equation}
where $P$ represents the number of pedestrians in a batch, and each pedestrian has $K_p$ images.

\subsection{Clothes Feature Minimization}\label{sec3.3}
Although IFP guides the model to directly extract cloth-agnostic features from raw images through cloth-agnostic text prompts, the lack of a mechanism to examine clothes features results in the possibility of clothes features being present in the extracted pedestrian features. Therefore, to further weaken the model's reliance on clothes clues from a negative perspective, a novel examine-disentanglement module, Clothes Feature Minimization (CFM), is proposed for iterative examination and disentangling clothes features from pedestrian features. This module mainly consists of two parts: Generating Clothes Text Prompts (GCTP) and Feature Disentanglement (FD).

\textbf{Generate Clothes Text Prompt}. 
Similar to the GCATP, we can obtain the clothes images $x^{c}_{i}$ that contain only the clothes information by utilizing the clothes-irrelevant areas mask $M_i^p$:
\begin{equation}
\begin{aligned}
x^{c}_{i} =   (I-M^{p}_{i}) \odot x_{i} +   M^{p}_{i}\odot 255
\end{aligned}.
\end{equation}
Then, we design an initial language description $T_{y_i^{c}}$ belonging to category $y_i^{c}$: "A photo of $[X]_1$ $[X]_2$ ... $[X]_M$ clothes". In the first stage of training, we freeze the CLIP's image encoder $E_{v}$ and text encoder $E_{t}$ and then obtain image features $\bm f^{c}_{{i}} = E_{v}(x^{c}_{i})$ and text features $\bm f^{t,c}_{{y_i^{c}}} = E_{t}(T_{y_i^{c}})$. To ensure that the learnable $T_{y_i^{c}}$ can accurately describe the clothes, we optimize $T_{y_i^{c}}$ using contrastive losses as follows:
\begin{equation}
	\begin{aligned}
	\mathcal{L}_{i2t}^{c} =  - \frac{1}{{{n_b}}}\sum\limits_{i = 1}^{{n_b}}  {\log \frac{{\exp \left( {s\left( {\bm f_{i}^c,\bm f_{y_i^c}^{t,c}} \right)} \right)}}{{\sum\nolimits_{j = 1}^{{n_b}} {\exp \left( {s\left( {\bm f_i^c,\bm f_{y_j^c}^{t,c}} \right)} \right)} }}}   
	\end{aligned},
\end{equation}

\begin{equation}
	\begin{aligned}
	\mathcal{L}_{t2i}^{c} =   - \frac{1}{{{n_b}}}\sum\limits_{i = 1}^{{n_b}}  {\log \frac{{\exp \left( {s\left( {\bm f_{i}^c,\bm f_{y_i^c}^{t,c}} \right)} \right)}}{{\sum\nolimits_{j = 1}^{{n_b}} {\exp \left( {s\left( {\bm f_j^c,\bm f_{y_i^c}^{t,c}} \right)} \right)} }}}  
	\end{aligned}.
\end{equation}

\textbf{Feature Disentanglement}. 
To examine clothes features contained in pedestrian feature $f_i$, we employ a learnable projection matrix $Proj_c\in\mathbb{R}^{C \times C}$ \cite{feng2023shape}, $C$ is the channel dimension of $f_i$. Then, the clothes feature $f_i^c$ can be obtained through $f_i^c=f_i \times Proj_c$. To ensure the discriminability of $f_i^c$, we align the text feature $f_i^{c,t} = E_t(T_{y_i^{c}})$ with the clothes feature $f_i^c$ by $\mathcal{L}_{i2tce}^{c}$, as follows:
\begin{equation}
	\begin{aligned}
	\mathcal{L}_{i2tce}^{c} =  - \frac{1}{{{n_b}}}\sum\limits_{i = 1}^{{n_b}}  {\log \frac{{\exp \left( {s\left( {\bm f_{i}^{c},\bm f^{t,c}_{{y_i^{c}}}} \right)} \right)}}{{\sum\nolimits_{y^{c} = 1}^{{K_c}} {\exp \left( {s\left( {\bm f_i^{c},\bm f^{t,c}_{{y^{c}}}} \right)} \right)} }}}   
	\end{aligned},
\end{equation}
where $K_c$ is the total number of the clothes category. It's worth noting that the loss $\mathcal{L}_{i2tce}^{c}$ only optimizes the weight of the projection matrix $Proj_c$. After obtaining the clothes features $f_i^c$, we optimize the $E_v$ to disentangle these clothes features by ensuring that the pedestrian features $f_i$ are dissimilar to the clothes features $f_i^c$:
\begin{equation}
	\begin{aligned}
	\mathcal{L}_{dis} =  - \frac{1}{{{n_b}}}\sum\limits_{i = 1}^{{n_b}} \frac{f_i \cdot f_i^c}{\|f_i\|_2 \cdot \|f_i^c\|_2}   
	\end{aligned}.
\end{equation}
In this process, the $\mathcal{L}_{i2tce}^{c}$ loss continually improves the ability of the projection matrix to extract clothes features. Subsequently, the $\mathcal{L}_{dis}$ loss optimizes the image encoder $E_v$ to ensure that the extracted pedestrian features do not contain the clothes features output by the projection matrix. These two loss, $\mathcal{L}_{i2tce}^{c}$ and $\mathcal{L}_{dis}$, alternate execution, continually examine and disentangle clothes features, thereby achieving the disentanglement of clothes features from pedestrian features, and finally obtaining the intrinsic pedestrian features.

\begin{table*}[!ht]\centering\small
\caption{Results of mAP and CMC (\%) obtained by our proposed method and the state-of-the-art Re-ID methods on PRCC, LTCC  and Vc-Clothes. ``R@1'' denote Rank-1. the ``\underline{ }'' and bold values indicate suboptimal and best results, respectively. }
\label{SOTA}
\begin{tabular}{m{3.5cm}<{\centering}
m{0.85cm}<{\centering}m{0.8cm}<{\centering}m{0.85cm}<{\centering}m{0.8cm}<{\centering}
m{0.85cm}<{\centering}m{0.8cm}<{\centering}m{0.85cm}<{\centering}m{0.8cm}<{\centering}
m{0.85cm}<{\centering}m{0.8cm}<{\centering}m{0.85cm}<{\centering}m{0.8cm}<{\centering}}

\toprule[0.8pt]
\multirow{3}*{Methods}  & \multicolumn{4}{c}{LTCC} & \multicolumn{4}{c}{PRCC} & \multicolumn{4}{c}{VC-Clothes}  \\ 
\cmidrule(lr){2-5} \cmidrule(lr){6-9} \cmidrule(lr){10-13}
& \multicolumn{2}{c}{General} & \multicolumn{2}{c}{Cloth-changing}
& \multicolumn{2}{c}{Same-clothes} & \multicolumn{2}{c}{Cloth-changing}
& \multicolumn{2}{c}{General} & \multicolumn{2}{c}{Cloth-changing}
\\ 
\cmidrule(lr){2-3} \cmidrule(lr){4-5} \cmidrule(lr){6-7} \cmidrule(lr){8-9} \cmidrule(lr){10-11} \cmidrule(lr){12-13}
    & {R@1}  & {mAP} & {R@1}  & {mAP} & {R@1}  & {mAP} & {R@1}  & {mAP}  & {R@1}  & {mAP} & {R@1}  & {mAP}         \\ \toprule[0.8pt]

  HACNN (CVPR18) \cite{li2018harmonious}   &60.2 &26.7 &21.6 &9.3 &82.5 &84.8 &21.8 &23.2 &68.6 &69.7 &49.6 &50.1  \\
  PCB (ECCV18) \cite{sun2018beyond}    &65.1 &30.6 &23.5 &10.0 &99.8 &97.0 &41.8 &38.7 &87.7 &74.6 &62.0 &62.2\\
  IANet (CVPR19) \cite{hou2019interaction}   &63.7 &31.0 &25.0 &12.6 &99.4 &98.3 &46.3 &45.9 &- &- &- &-\\
  OSNet  (ICCV19) \cite{zhou2019omni}                                     &67.9 &32.1 &23.9 &10.8 &- &- &- &- &- &- &- &- \\
  ISP (ECCV20) \cite{zhu2020identity}  &66.3 &29.6 &27.8 &11.9 &92.8 &- &36.6 &- &94.5 &94.7 &72.0 &72.1\\ 
  \hline

  FSAM (CVPR21) \cite{hong2021fine}   &73.2 &35.4 &38.5 &16.2 &- &- &- &- &94.7 &\textbf{94.8} &78.6 &78.9\\
  RCSANet (ICCV21) \cite{huang2021clothing} &- &- &- &- &100 &97.2 &50.2 &48.6  &- &- &- &-\\
    GI-ReID (CVPR22) \cite{jin2022cloth} &63.2 &29.4 &23.7 &10.4 &- &- &- &- &- &- &64.5 &57.8\\
  CAL (CVPR22) \cite{gu2022clothes}    &74.2 &40.8 &40.1 &18.0 &100 &99.8 &55.2 &55.8 &92.9 &87.2 &81.4 &81.7\\
    UCAD (IJCAI22) \cite{yan2022weakening} &74.4 &34.8 &32.5 &15.1 &96.5 &- &45.3 &- &- &- &- &- \\
  M2Net (ACM MM22) \cite{liu2022long} &- &- &- &- &99.5 &99.1 &59.3 &57.7 &- &- &- &-\\
ACID(TIP23) \cite{yang2023win}   &65.1 &30.6 &29.1 &14.5 &99.1 &99.0 &55.4 &\textbf{66.1} &\underline{95.1} &\underline{94.7} &\underline{84.3} &74.2\\
3DInvarReID(ICCV23) \cite{liu2023learning}   &- &- &40.9 &18.9 &- &- &56.5 &57.2 &- &- &- &-\\
  AIM (CVPR23) \cite{yang2023good}     &\textbf{76.3} &\underline{41.1} &40.6 &19.1 &\textbf{100} &\textbf{99.9} &57.9 &58.3 &- &- &- &-\\
  MBUNet(TIP23) \cite{zhang2023multi}   &67.6 &34.8 &40.3 &15.0 &\textbf{100} &99.6 &\underline{68.7} &\underline{65.2} &\textbf{95.4} &94.3 &82.7 &70.3\\
MCSC(TIP24) \cite{huang2024meta}     &73.9 &40.2 &\underline{42.2} &\underline{19.4} &99.8 &\underline{99.8} &57.8 &57.3 &93.2 &87.9 &83.3 &\underline{83.2}\\
 \hline
   \textbf{CCAF(our)}  &\underline{75.3} &\textbf{41.3} &\textbf{42.9} &\textbf{20.1} &\underline{99.9} &98.4 &\textbf{70.4} &63.7 &\textbf{95.4} &90.9 &\textbf{88.6} &\textbf{87.2}   \\ \toprule[0.8pt]
\end{tabular}
\end{table*}

\subsection{Training and Inference}\label{sec3.4}
\noindent\textbf{Training}. 
The entire training process is divided into two stages. In the first stage, we train the model to generate text prompts for the clothes images and the shielding images by contrastive loss. The total loss is as follows:
\begin{equation}
	\begin{aligned}
	\mathcal{L}_{stage1} =  \mathcal{L}_{i2t}  + \mathcal{L}_{t2i} +  \mathcal{L}_{i2t}^{c}  + \mathcal{L}_{t2i}^{c}
	\end{aligned}.
\end{equation}
In the second stage of training, we achieve image-text alignment and image-image alignment by applying the losses $\mathcal{L}_{i2tce} $ and $\mathcal{L}_{con}$, respectively, to guide the model to focus on the cloth-agnostic features of pedestrians. Subsequently, we further apply losses $ \mathcal{L}_{i2tce}^{c}$ and $\mathcal{L}_{dis}$ to examine and disentangle clothes features from negative aspects. The total loss is as follows:
\begin{equation}
	\begin{aligned}
	\mathcal{L}_{stage2} &=  \mathcal{L}_{id} + \mathcal{L}_{i2tce}  +  \mathcal{L}_{id}^{s}  + \mathcal{L}_{i2tce}^{s}\\  & + \mathcal{L}_{i2tce}^{c} +  \lambda_1 \mathcal{L}_{con} 
+  \lambda_2 \mathcal{L}_{dis}
	\end{aligned},
\end{equation}
where $\lambda_1$ and $\lambda_2$ are hyper-parameters used to balance loss items.

\noindent\textbf{Inference}. Note that only the $E_v$ is used for inference, therefore, the IFP and CFM proposed do not introduce additional computational complexity during the inference stage. During inference, we first extract the features of pedestrian images from the query set and gallery set, and then calculate the cosine similarity between them.

\section{Experiments}

\subsection{Datasets and Evaluation Metrics}
\textbf{PRCC} \cite{yang2019person} comprises 33,698 images from 221 identities captured across 3 cameras. Individuals wear the same clothes in the A and B cameras, whereas they wear distinct clothes in the C camera. 

\noindent\textbf{LTCC} \cite{qian2020long} comprises 17,138 images of 152 different identities, all captured from 12 camera perspectives. Among them, the dataset contains 478 unique sets of clothes.

\noindent\textbf{VC-Clothes} \cite{wan2020person} consists of 512 pedestrians covering 4 scenes, with an average of 9 images per scene, totaling 19,060 pedestrian images. The image was collected from Grand Theft Auto V (GTA5). 

\noindent\textbf{Deepchange} \cite{xu2023deepchange} is a large-scale person re-identification dataset spanning 2 years, comprising over 170,000 images of 1,121 pedestrians captured by 17 cameras. Among them, 450 pedestrians were used for training, 150 for the validation set, and 521 for testing.

\noindent\textbf{Evaluation Metrics}. We used Cumulative Matching Characteristics (CMC) and Mean Average Precision (mAP) as metrics to evaluate the retrieval performance of the proposed CCAF.

\vspace{-6.5pt}
\subsection{Implementation Details}
We implemented the proposed CCAF on the PyTorch deep learning framework, and all experiments were conducted on a single A100 GPU. We used CLIP's image encoder (ViT-B-16) as the backbone network, the input images were uniformly resized to $256 \times 128$, and common data augmentation strategies were applied to enhance the input images, such as random horizontal flips, padding, random crop, and random erasing \cite{zhong2020random}. 
In the first stage, we trained two prompt learners $T_{y^{id}}$ and $T_{y^{c}}$ with an initial learning rate set to 3.5e-4 for 120 epochs and batch size set to 64. 
In the second stage, we fine-tuned two image encoders ($E_v$, $E_v^{s}$) and one projection layer $Proj_c$ for 40 epochs, with an initial learning rate set to 5e-6. 
Both stages adopt the Adam optimizer with the cosine learning rate decay strategy\cite{he2021transreid} to adjust the learning rate dynamically. The batch size was set to 64, consisting of 16 pedestrians with 4 images each.
The hyper-parameters $\lambda_1$ and $\lambda_2$ were set to 0.1 and 1, respectively.

\vspace{-4pt}
\subsection{Comparison with State-of-the-Art Methods}
In this section, we conducted comparative experiments with other state-of-the-art methods on the PRCC, LTCC, VC-Clothes, and Deepchange datasets. Specifically, the methods we compared include short-term person ReID methods: HACNN \cite{li2018harmonious}, PCB \cite{sun2018beyond}, IANet \cite{hou2019interaction}, ISP \cite{zhu2020identity}, and CC-ReID methods: FSAM \cite{hong2021fine}, RCSANet \cite{huang2021clothing}, CAL \cite{gu2022clothes}, GI-ReID \cite{jin2022cloth}, M2Net \cite{liu2022long}, AIM \cite{yang2023good}, MBUNet \cite{zhang2023multi}, ACID \cite{yang2023win}, 3DInvarReID \cite{liu2023learning}  and MCSC \cite{huang2024meta}.
\begin{table}[!ht]\centering\small
 \caption{Results of mAP and CMC (\%) obtained by our proposed method and the state-of-the-art Re-ID methods on DeepChange. ``R@1'' denote Rank-1. }
\label{SOTA2}

\begin{tabular}{m{3.5cm}<{\centering}m{2.4cm}<{\centering}m{1.5cm}}
\toprule[0.8pt]

\multirow{2}*{Methods}  & \multicolumn{2}{c}{DeepChange}   \\ \cmidrule(lr){2-3} 
& {R@1}  & {mAP}\\ \toprule[0.8pt]
MGN (ACM MM18)  \cite{wang2018learning}    &25.4 &9.8   \\
ABD-Net (ICCV19) \cite{chen2019abd}    &24.2 &8.5 \\
OSNet (ICCV19)  \cite{zhou2019omni}    &39.7 &10.3 \\
RGA-SC (CVPR20)  \cite{zhang2020relation}   &28.9 &8.6 \\
ReIDCaps (TCSVT20) \cite{huang2019beyond} &39.5 &11.3\\
TransreID (ICCV21) \cite{he2021transreid} &35.9 &14.4 \\
CAL (CVPR22) \cite{gu2022clothes}       &54.0 &19.0 \\
MCSC (TIP24) \cite{huang2024meta}        &\underline{56.9} &\textbf{21.5} \\
 \hline
\textbf{CCAF(our)}  &\textbf{59.6} &\underline{21.0}    \\ \toprule[0.8pt]
\end{tabular}
\end{table}
\begin{table}[!ht]\small
 \centering {\caption{Ablation studies of the proposed CCAF. ``B'': Baseline. ``I2T'': Image-Text alignment. ``I2I'': Image-Image alignment. ``CFM'': Clothes Feature Minimization.}\label{Tab:5}
\begin{tabular}{m{0.45cm}<{\centering}m{0.45cm}<{\centering}m{0.45cm}<{\centering}m{0.45cm}<{\centering}m{0.7cm}<{\centering}m{0.7cm}<{\centering}m{0.7cm}<{\centering}m{0.7cm}<{\centering}}
\toprule[0.8pt]
  \multicolumn{4}{c}{Component}  & \multicolumn{2}{c}{LTCC}  & \multicolumn{2}{c}{PRCC} \\ 
\cmidrule(lr){1-4} \cmidrule(lr){5-6} \cmidrule(lr){7-8} 
  B &I2T & I2I & CFM                              & R@1  & mAP   & R@1  & mAP \\ 
\toprule[0.8pt]
 $\CIRCLE$ &\Circle   &\Circle    &\Circle      & 35.2 & 17.4  & 62.9 & 59.1 \\
 $\CIRCLE$ &$\CIRCLE$ &\Circle    &\Circle      & 39.0 & 18.6  & 66.7 & 61.3 \\
 $\CIRCLE$ &$\CIRCLE$ &$\CIRCLE$  &\Circle      & 40.3 & 19.8  & 69.6 & 63.6 \\
 $\CIRCLE$ &$\CIRCLE$ &$\CIRCLE$  &$\CIRCLE$    & 42.9 & 20.1  & 70.4 & 63.7 \\
\toprule[0.8pt]
\end{tabular}}
\end{table}
On the LTCC dataset, our method achieves a Rank-1/mAP of 75.3\%/41.3\% in the general setting and 42.9\%/20.1\% in the cloth-changing setting. Specifically, focusing on the cloth-changing setting, our method outperforms the second-best method, MCSC, by 0.7\%/0.7\%. On the PRCC dataset, in the same-clothes setting, our CCAF achieves a Rank-1/mAP of 99.9\%/98.4\%, and in the cloth-changing setting, it achieves 70.4\%/63.7\%, surpassing the second-best method MBUNet by 1.7\% in Rank-1. On the larger-scale virtual dataset VC-Clothes, our CCAF achieves a Rank-1/mAP of 95.4\%/90.9\% in the general setting and surpasses the second-best method, MCSC, by 5.3\%/4.0\% in the cloth-changing setting. To verify the performance of our CCAF in more complex scenarios, we also conducted experiments on the large-scale real-world dataset DeepChange, where our method surpasses the second-best method, MCSC, by 2.7\% in Rank-1. 
The results on small-scale, large-scale, and synthetic datasets (especially in the cloth-changing setting) demonstrate the effectiveness of the proposed CCAF. This is attributed to CCAF's efficient utilization of CLIP's cross-modal knowledge, enabling the extraction of high-level semantic features that are independent of clothes.

\vspace{-4pt}
\subsection{Ablation Study}
To evaluate the effectiveness of each component of our CCAF, we conducted a series of ablation experiments on the PRCC and LTCC datasets in the cloth-changing setting, as shown in Tab. \ref{Tab:5}. The baseline method trained the $E_v$ using only $\mathcal{L}_{ce}$ and $\mathcal{L}_{tri}$ losses. The components evaluated include I2T and I2T of Invariant Feature Prompting (IFP), and Clothes Feature Minimization (CFM).

\noindent\textbf{Effectiveness of I2T.} 
To validate the effectiveness of I2T, we integrated it into the baseline method, resulting in ``B+I2I''. Compared to the baseline, on the LTCC and PRCC datasets, Rank-1 (mAP) improved by 3.8\% (1.2\%) and 3.8\% (2.2\%), respectively. These performance improvements indicate that the cloth-agnostic text prompts generated from shielding images effectively describe the fine-grained semantic features of pedestrians. Furthermore, during alignment, the model is forced to focus more on the cloth-agnostic features and avoid overly focusing on clothes-related clues.
\begin{table}[!ht]\small
 \centering {\caption{Ablation studies of the proposed loss. ``w/o''denote without , ``r/p $\mathcal{L}_{i2tce}^{c}$'' denote replace $\mathcal{L}_{i2tce}^{c}$ by the cross entropy loss supervised by one-hot label}\label{Tab:text prompt}
\begin{tabular}{m{0.8cm}<{\centering}m{0.8cm}<{\centering}m{0.8cm}<{\centering}m{0.7cm}<{\centering}m{0.7cm}<{\centering}m{0.7cm}<{\centering}m{0.7cm}<{\centering}}
\toprule[0.8pt]
  \multicolumn{3}{c}{Loss}  & \multicolumn{2}{c}{LTCC}  & \multicolumn{2}{c}{PRCC} \\ 
\cmidrule(lr){1-3} \cmidrule(lr){4-5} \cmidrule(lr){6-7} 
w/o $\mathcal{L}_{i2tce}$ & w/o $\mathcal{L}_{i2tce}^{s}$ &r/p $\mathcal{L}_{i2tce}^{c}$                              & R@1  & mAP   & R@1  & mAP \\ 
\toprule[0.8pt]


 \Circle  &\Circle   &\Circle     & 42.9 & 20.1  & 70.4 & 63.7 \\
  \Circle  &\Circle  &$\CIRCLE$      & 41.6 \textcolor{red}{\scalebox{0.8}{$\downarrow$}} & 19.6 \textcolor{red}{\scalebox{0.8}{$\downarrow$}} & 70.1 \textcolor{red}{\scalebox{0.8}{$\downarrow$}} & 63.4 \textcolor{red}{\scalebox{0.8}{$\downarrow$}} \\
   \Circle  &$\CIRCLE$ &$\CIRCLE$         & 41.1 \textcolor{red}{\scalebox{0.8}{$\downarrow$}} & 19.2 \textcolor{red}{\scalebox{0.8}{$\downarrow$}} & 68.9 \textcolor{red}{\scalebox{0.8}{$\downarrow$}} & 62.9 \textcolor{red}{\scalebox{0.8}{$\downarrow$}} \\
    $\CIRCLE$ &$\CIRCLE$ &$\CIRCLE$          & 35.7 \textcolor{red}{\scalebox{0.8}{$\downarrow$}} & 17.6 \textcolor{red}{\scalebox{0.8}{$\downarrow$}} & 67.0 \textcolor{red}{\scalebox{0.8}{$\downarrow$}} & 59.6 \textcolor{red}{\scalebox{0.8}{$\downarrow$}} \\
\toprule[0.8pt]
\end{tabular}}
\end{table}

\noindent\textbf{Effectiveness of I2I.} 
To evaluate the effectiveness of I2I, we integrated it into ``B+I2T'' to obtain ``B+I2T+I2I''. Compared to the ``B+I2'', on the LTCC dataset, ``B+I2T+I2I'' achieved improvements of 1.3\% and 1.2\% in Rank-1 and mAP, respectively. On the PRCC dataset, ``B+I2T+I2I'' improved Rank-1 and mAP by 2.9\% and 2.3\%, respectively. These results indicate that I2I can further restrict the model from overly focusing on clothes information by achieving alignment at the feature level within the same modality.

\noindent\textbf{Effectiveness of CFM.} 
To validate the effectiveness of CFM, we integrated it into ``B+I2T+I2I'' to obtain CCAF. Compared to ``B+I2T+I2I'', our CCAF achieved improvements of 2.6\% (0.8\%) and 0.3\% (0.1\%) in Rank-1 and mAP, respectively, on the LTCC (PRCC) dataset. This indicates that CFM can examine and disentangle clothes features at the feature level, enabling the model to extract cloth-agnostic features from the raw images from the negative perspective. 

\noindent\textbf{Effectiveness of text prompt.} 
To validate the effectiveness of the text prompts, as shown in Tab. \ref{Tab:text prompt}, we first replaced the clothes prompts with one-hot labels in CFM, further removed the supervision of text prompts on the shielding stream, and finally removed the supervision of text prompts on the raw stream. It can be visually observed that the text prompt supervision plays a role in the proposed AACF.

\begin{figure}[t!]
\centering
\includegraphics[width=8cm,keepaspectratio=true]{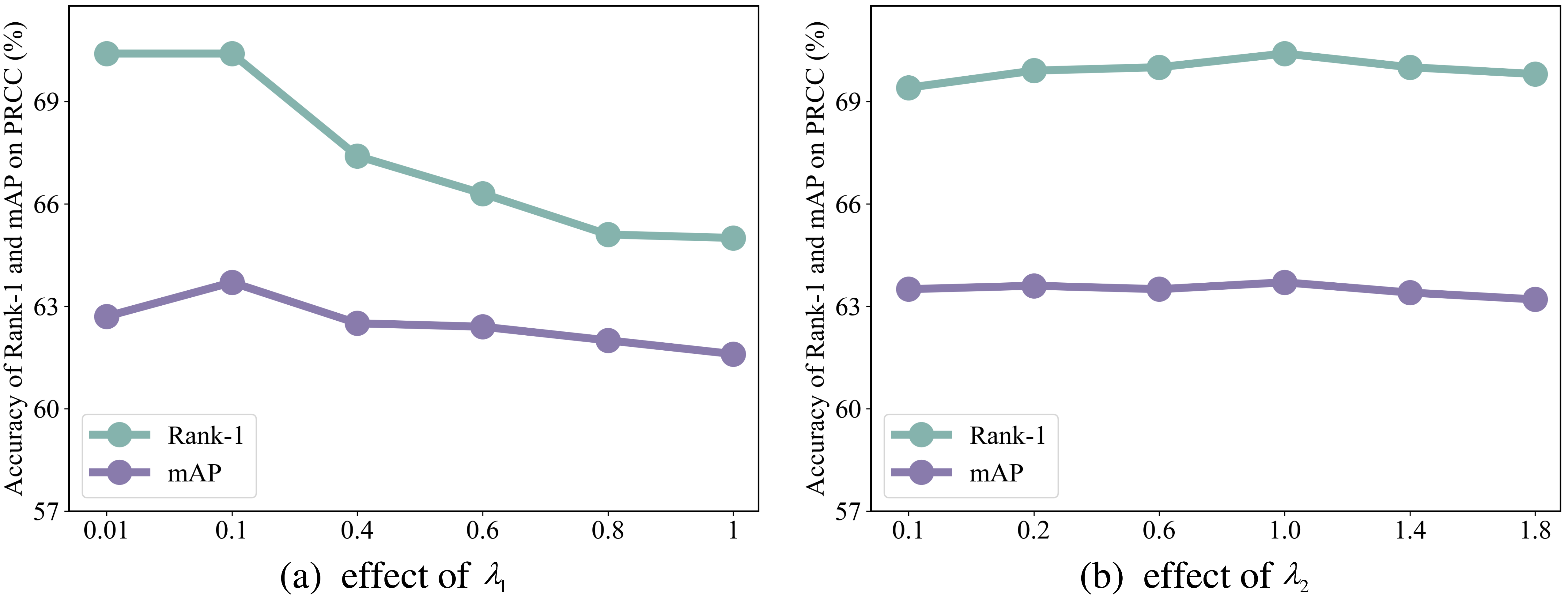}
\caption{Results of Rank-1 and mAP with different values of $\lambda_{1}$ and $\lambda_{2}$ on PRCC dataset. }
\label{fig:param}
\end{figure}
\noindent\textbf{Effect of hyper-parameters.} 
The proposed method involves two hyperparameters \(\lambda_1\) and \(\lambda_2\), which are used to balance the weights of loss terms $\mathcal{L}_{con}$ and $\mathcal{L}_{dis}$, respectively. To explore the impact of different values of hyper-parameters on performance, we conducted a parameter analysis experiment on PRCC dataset by fixing one parameter and adjusting the other. As shown in Fig. \ref{fig:param}, the model achieves the best performance when \(\lambda_1\) is set to 0.1 and \(\lambda_2\) is set to 1.

\begin{figure}[t!]
\centering
\includegraphics[width=8.5cm,keepaspectratio=true]{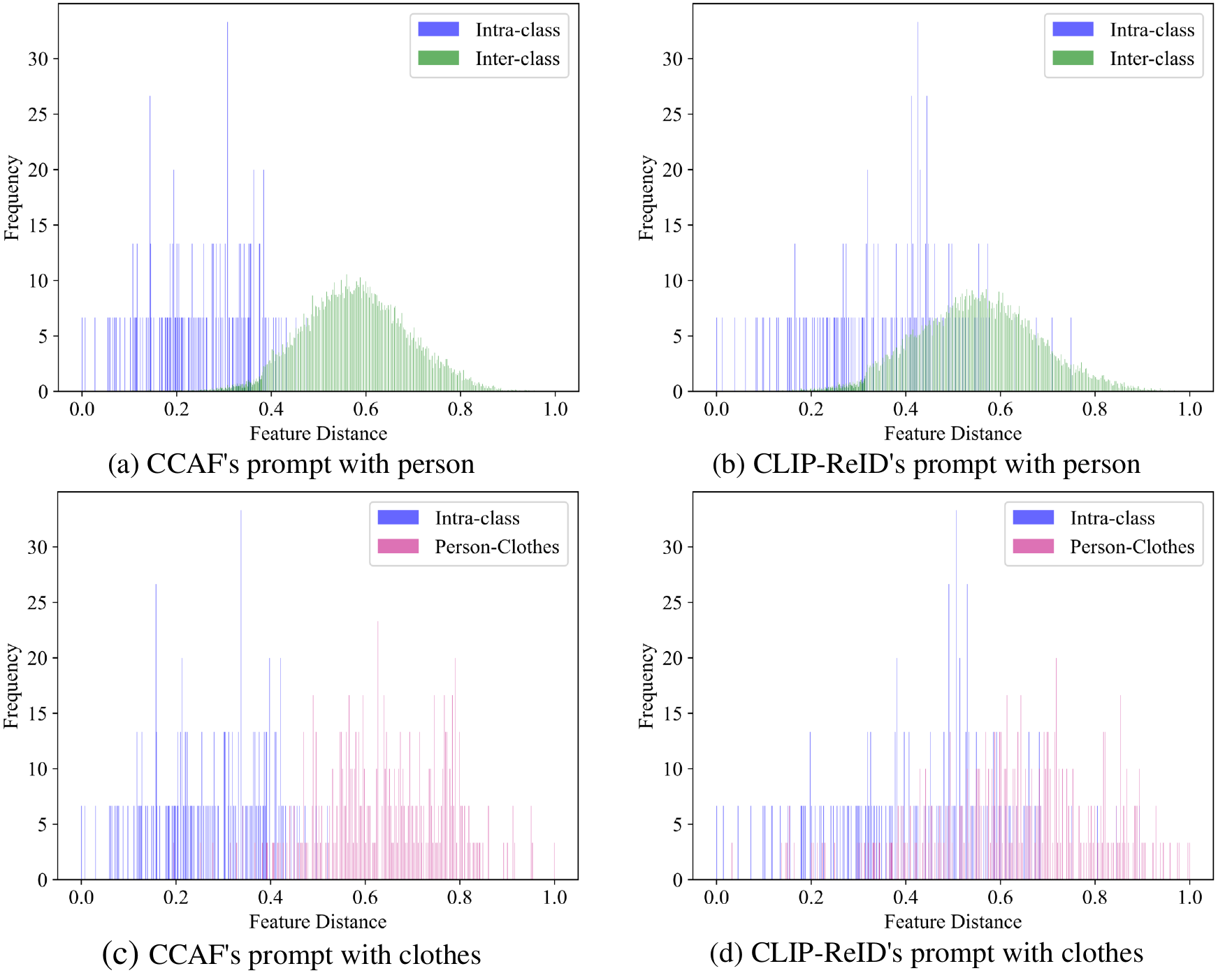}
\caption{The distance distribution of features between pedestrian's prompt and shielding (clothes) images on PRCC dataset.}
\label{prompt}
\end{figure}

\begin{figure}[t!]
\centering
\includegraphics[width=8.5cm,keepaspectratio=true]{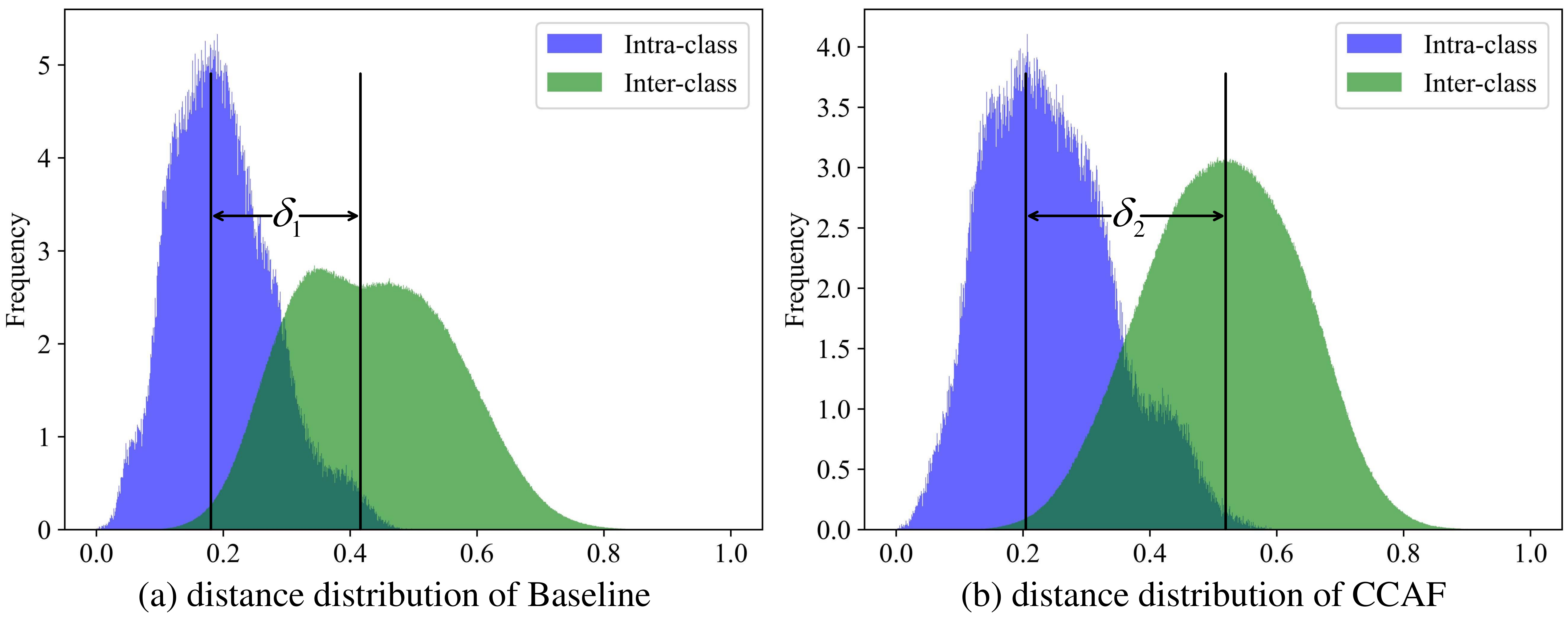}
\caption{The distance distribution of features between intra-class and inter-class samples on PRCC dataset. }
\label{distance}
\end{figure}

\begin{figure}[t!]
\centering
\includegraphics[width=8.5cm,keepaspectratio=true]{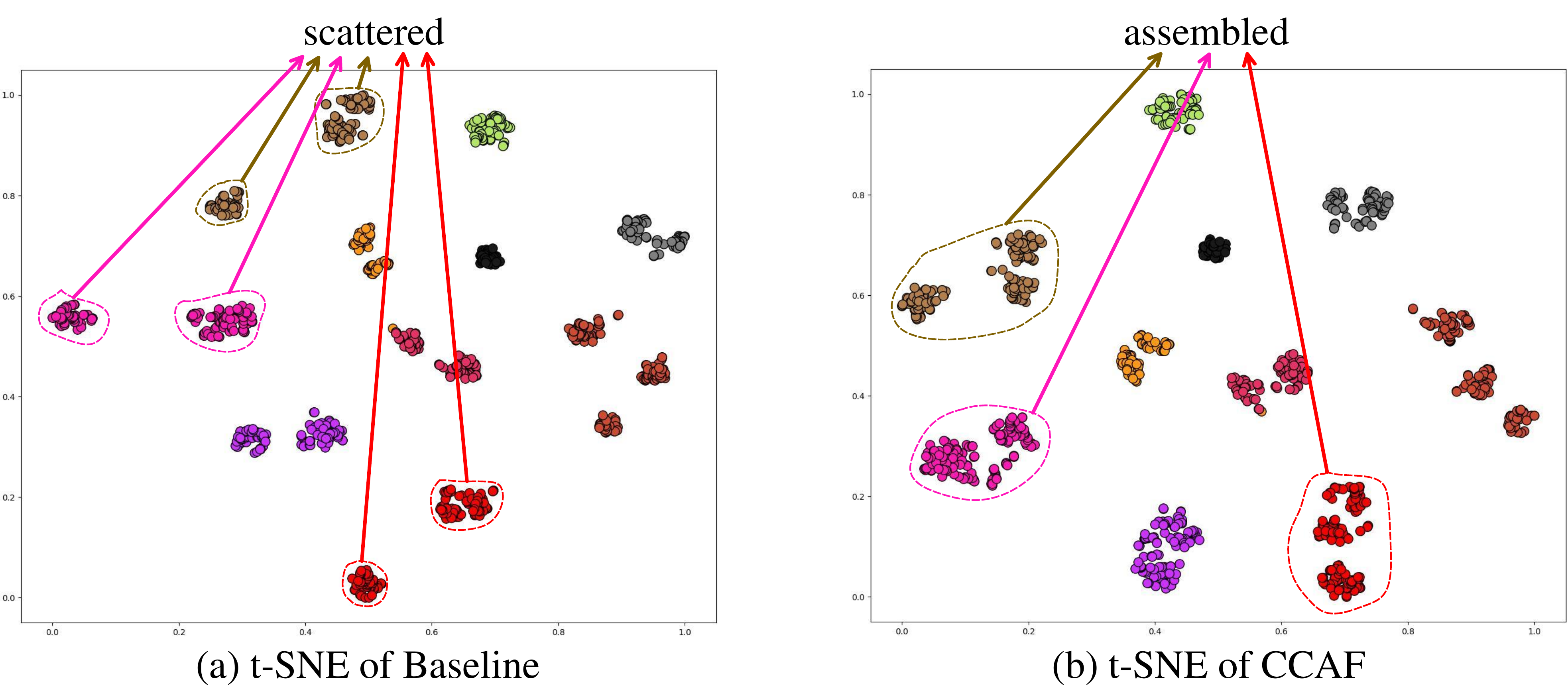}
\caption{Visualization of 10 randomly selected pedestrians from the PRCC dataset.  }
\label{tsne}
\end{figure}
\begin{figure}[t!]
\centering
\includegraphics[width=8.5cm,keepaspectratio=true]{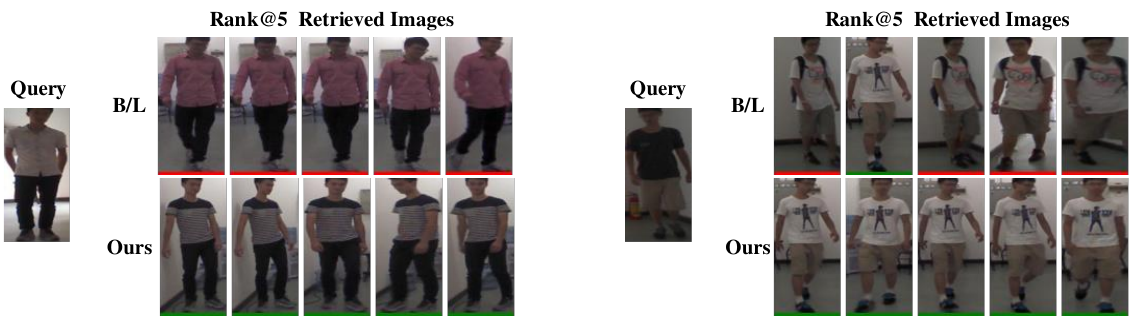}
\caption{Visualization of pedestrian search results, where B/L represents the baseline method. }
\label{retrieval}
\end{figure}

\section{Visualization}
\noindent\textbf{Features Distribution Analysis.}
 To investigate whether the generated text prompts are identity-related and not related to clothes, we visualized the distance distributions of features between the pedestrian's text prompts of CCAF (CLIP-ReID) and the shielding images. As shown in Fig. \ref{prompt} (a) (b), it is visually evident that our prompts are more discriminative, while CLIP-ReID's prompts lack discriminative power due to describing clothes. Furthermore, as shown in Fig. \ref{prompt} (c) (d), we visualized the distance distributions between the text prompts of CCAF (CLIP-ReID) and the corresponding clothes images, demonstrating that our prompts are unrelated to clothes, while CLIP-ReID's prompts are associated with clothes.

Furthermore, we visualize the feature distance distributions of the intra-class and inter-class samples on the PRCC dataset. Fig. \ref{distance} (a) shows the distance distributions of the intra-class and inter-class for the baseline method, while Fig. \ref{distance} (b) shows the distance distributions of intra-class and inter-class for our CCAF. It can be intuitively observed that after applying our method, the mean distances (indicated by the vertical lines) between the intra-class and inter-class are increased (i.e., $\delta_2 > \delta_1$), indicating an increase in the mean distances between the intra-class and inter-class, enhancing the discriminability between different classes. Therefore, the proposed CCAF can effectively learn fine-grained cloth-agnostic features.

 Meanwhile, we also visualized the feature distributions using t-SNE \cite{van2008visualizing}. In Fig.\ref{tsne} (a), we can see that different feature embeddings of the same pedestrian are separated. But, in Fig.\ref{tsne} (b), the proposed CCAF effectively aggregates these feature embeddings belonging to the same pedestrians, thus effectively alleviating the negative impact of clothes variations.

\noindent\textbf{Retrieval Results Analysis.}   
To further validate the effectiveness of the proposed CCAF, we also visualize the pedestrian retrieval results of the baseline method and our CCAF on the PRCC dataset in Fig. \ref{retrieval}. Specifically, given a query pedestrian image, we retrieve the top 5 pedestrian images with the highest similarity. The green bar at the bottom indicates the correctly retrieved images of the same identity, while the red bar indicates at the bottom indicates the incorrectly retrieved images. From the retrieval results, it can be seen that compared to the baseline method, our proposed CCAF can retrieve images of pedestrians wearing different clothes, while the baseline retrieves incorrect pedestrian images.

\section{Conclusion}
In this paper, we propose a novel framework called CCAF, designed to leverage CLIP's knowledge to learn fine-grained semantic features unrelated to clothes, thereby enhancing CC-ReID. Specifically, we generate cloth-agnostic text prompts and clothes text prompts by exploring CLIP's knowledge. With the help of these two types of text prompts, the proposed CCAF not only emphasizes features that are unrelated to clothes but relevant to identity from the positive aspect but also examines the presence of clothes features in pedestrian features from a negative aspect and disentangles clothes features from pedestrian features, thereby improving performance. The superior performance observed across four publicly available CC-ReID benchmarks demonstrates the effectiveness and superiority of the proposed CCAF.

\bibliographystyle{ACM-Reference-Format}
\bibliography{sample-base}










\end{document}